# Systematic vs. Non-systematic Algorithms for Solving the MPE Task

Radu Marinescu        Kalev Kask        Rina Dechter
School of Information and Computer Science
University of California, Irvine, CA 92697-3425
{radum,kkask,dechter}@ics.uci.edu

## Abstract

The paper explores the power of two systematic Branch and Bound search algorithms that exploit partition-based heuristics, BBBT (a new algorithm for which the heuristic information is constructed during search and allows dynamic variable/value ordering) and its predecessor BBMB (for which the heuristic information is pre-compiled) and compares them against a number of popular local search algorithms for the MPE problem as well as against the recently popular iterative belief propagation algorithms. We show empirically that the new Branch and Bound algorithm, BBBT demonstrates tremendous pruning of the search space far beyond its predecessor, BBMB which translates to impressive time saving for some classes of problems. Second, when viewed as approximation schemes, BBBT/BBMB together are highly competitive with the best known SLS algorithms and are superior, especially when the domain sizes increase beyond 2. The results also show that the class of belief propagation algorithms can outperform SLS, but they are quite inferior to BBMB/BBBT. As far as we know, BBBT/BBMB are currently among the best performing algorithms for solving the MPE task.

## 1 INTRODUCTION

The paper presents an extensive empirical study of highly competitive approaches for solving the Most Probable Explanation (MPE) task in Bayesian networks introduced in recent years. We compare two Branch-and-Bound(BnB) algorithms that exploit bounded inference for heuristic guidance on the one hand, against incomplete approximation algorithms, such as stochastic local search, which have proven to be powerful for solving constraint satisfaction (CSP) and boolean satisfiability (SAT) problems in recent years, on the other. We also compare against the class of generalized iterative belief propagation adapted for the MPE task.

Our Branch-and-Bound algorithms are based on partitioning based approximation of inference, called *mini-bucket elimination* (MBE($i$)) first introduced in [Dechter and Rish 1997]). The mini-bucket scheme approximates variable elimination algorithms. Rather than computing and recording functions on many variables, as is often required by variable elimination, the mini-bucket scheme partitions function computations into subsets of bounded number of variables, $i$, (the so called $i$-bound), and records several smaller functions instead. It can be shown that it outputs an upper bound (resp., lower bound) on the desired optimal value for a maximization (resp., minimization) task. This is a flexible scheme that can tradeoff complexity for accuracy; as the $i$-bound increases both the computational complexity (which is exp($i$)) and the accuracy increase (for details see [Dechter and Rish 1997, Dechter and Rish 2003]).

It was subsequently shown in [Kask and Dechter 2001] that the functions generated by MBE($i$) can be used to create heuristic functions that guide search. These heuristics have varying strengths depending on the mini-bucket's $i$-bound, allowing a controlled tradeoff between pre-processing (for heuristics generation) and search. The resulting Branch and Bound with Mini-Bucket heuristic BBMB($i$), was evaluated extensively for probabilistic and deterministic optimization tasks. Results show that the scheme overcomes partially the memory explosion of bucket-elimination allowing a gradual tradeoff of space for time, and of time for accuracy.

More recently a new algorithm called BBBT($i$) [Dechter *et al.*2001] was introduced that takes the idea of partition-based heuristics one step further. It explores the feasibility of generating partition-based heuristics *during search*, rather than in a *pre-processing manner*. This allows dynamic variable ordering - a feature that can have tremendous effect on search. The dynamic generation of these heuristics is facilitated by a recent extension of mini-bucket elimination to mini-bucket *tree* elimination (MBTE), a partition-based approximation defined over cluster-trees described in [Dechter et al.2001]. MBTE outputs multiple (lower or upper) bounds for each possible variable and value extension at once, which is much faster than running MBE $n$ times, one for each variable, to generate the same result.

This yields algorithm BBBT($i$) that applies the MBTE($i$) heuristic computation at each node of the search tree. Clearly, the algorithm has a much higher time overhead compared with BBMB($i$) for the same $i$-bound, but it can prune the search space much more effectively, hopefully yielding overall superior performance for some classes of hard problems. Preliminary tests of the algorithms for the MAX-CSP (finding an assignment to a constraint problem that satisfies a maximum number of constraints) task showed that on a class of hard enough problems BBBT($i$) with the smallest $i$-bound ($i$=2) is cost-effective [Dechter et al.2001].

Stochastic Local Search (SLS) is a class of incomplete approximation algorithms which, unlike complete algorithms, are not guaranteed to find an optimal solution, but as shown during the last decade, are often far superior to complete systematic algorithm on CSP and SAT problems. In this paper we will compare a number of best-known SLS algorithms for solving the MPE problem against BBMB/BBBT. Some of these SLS algorithms are applied directly on the Bayesian network, some translate the problem into a weighted SAT problem first and then apply a MAX-wSAT algorithm.

A third class of algorithms are iterative join-graph propagation (IJGP($i$)) that applies Pearl's belief propagation algorithm to loopy join-graphs of the belief network [Dechter et al.2002].

We experiment with random uniform, Noisy-OR, NxN grid and random coding problems, as well as a number of real world benchmarks. Our results show that BBMB and BBBT do not dominate one another. While BBBT can sometimes significantly improve over BBMB, in many other instances its (quite significant) pruning power does not outweigh its time overhead. Both algorithms are powerful in different cases. In general when large $i$-bounds are effective BBMB is more powerful, however when space is at issue BBBT with small $i$-bound is often more powerful. More significantly, we show that SLS algorithms are overall inferior to BBBT/BBMB, except when the domain size is small. This is unlike what we often see in the case of CSP/SAT problems, especially in the context of randomly generated instances. The superiority of BBBT/BBMB is especially significant because unlike local search they can prove optimality if given enough time. Finally, we demonstrate that generalized belief propagation IJGP($i$) algorithms are often superior to the SLS class as well.

In Section 2 we present background, definitions and describe relevant recent work on mini-bucket-tree elimination underlying the BBBT algorithm. Section 3 presents an overview of BBBT, contrasted with BBMB. Section 4 provides an overview of the SLS algorithms used in our study and the iterative join-graph propagation algorithms. In Section 5 we provide our experimental results, while Section 6 concludes.

## 2 BACKGROUND

### 2.1 PRELIMINARIES

**Belief networks.** A *belief network* is a quadruple $BN = <X, D, G, P>$ (also abbreviated $<G, P>$) where $X = \{X_1, ..., X_n\}$ is a set of random variables, $D = \{D_1, ..., D_n\}$ is the set of the corresponding domains, $G$ is a directed acyclic graph over $X$ and $P = \{P_1, ..., P_n\}$, where $P_i = P(X_i|pa_i)$ ($pa_i$ are the parents of $X_i$ in $G$) denote conditional probability tables (CPTs). Given a function $f$, we denote by $scope(f)$ the set of arguments of function $f$. The family of $X_i$, $F_i$, includes $X_i$ and its parent variables.

The most common automated reasoning tasks in Bayesian networks are *belief updating, most probable explanation (MPE)* and *maximum aposteriory probability (MAP)*. In this paper we focus on the MPE task. It appears in applications such as diagnosis, abduction and explanation. For example, given data on clinical findings, MPE can postulate on a patient's probable affliction. In decoding, the task is to identify the most likely input message transmitted over a noisy channel, given the observed output.

DEFINITION 2.1 (MPE) *The most probable explanation problem is to find a most probably complete assignment that is consistent with the evidence, namely, to find an assignment $(x_1^o, ..., x_n^o)$ such that $P(x_1^o, ..., x_n^o) = max_{x_1,...,x_n} \prod_{k=1}^{n} P(x_k, e|x_{pa_k})$*

**Singleton-optimality task.** In addition to finding the global optimum (MPE), of particular interest to us is the special case of finding, for each assignment $X_i = x_i$, the highest probability of the complete assignment that agrees with $X_i = x_i$. Formally, we want to compute $z(X_i) = max_{X-\{X_i\}}(\prod_{k=1}^{n} P_k)$, for each variable $X_i$.

The common exact algorithms for Bayesian inference are join-tree clustering defined over tree-decompositions [Lauritzen and Spiegelhalter 1988] and variable elimination algorithms [Dechter 1999]. The variant we use was presented recently for constraint problems.

DEFINITION 2.2 (cluster-tree decompositions) *[Gottlob et al.1999] Let $BN = <X, D, G, P>$ be a belief network. A cluster-tree decomposition for $BN$ is a triple $D = <T, \chi, \psi>$, where $T = (V, E)$ is a tree, and $\chi$ and $\psi$ are labeling functions which associate with each vertex $v \in V$ two sets, $\chi(v) \subseteq X$ and $\psi(v) \subseteq P$.*

1. *For each function $P_i \in P$, there is exactly one vertex $v \in V$ such that $p_i \in \psi(v)$, and $scope(p_i) \subseteq \chi(v)$.*

---

**Procedure CTE**
**Input:** A Bayesian network $BN$, a tree-decomposition $< T, \chi, \psi >$.
**Output:** A set of functions $z_i$ as a solution to the singleton-optimality task.
**Repeat**
  1. Select an edge $(u, v)$ such that $m_{(u,v)}$ has not been computed and $u$ has received messages from all adjacent vertices other than $v$.
  2. $m_{(u,v)} \leftarrow \max_{elim(u,v)} \prod_{g \in cluster(u), g \neq m_{(v,u)}} g$ (where $cluster(u) = \psi(u) \cup \{m_{(w,u)}|(w,u) \in T\}$).
**Until** all messages have been computed.
**Return** for each $i$, $z(X_i) = \max_{\chi(u) - X_i} \prod_{g \in cluster(u)} g$, such that $X_i \in cluster(u)$.

---

Figure 1: Algorithm cluster-tree elimination (CTE) for singleton-optimality task.

2. *For each variable $X_i \in X$, the set $\{v \in V | X_i \in \chi(v)\}$ induces a connected subtree of $T$. The connectedness requirement is also called the running intersection property.*

Let $(u, v)$ be an edge of a tree-decomposition, the separator *of $u$ and $v$ is defined as* $sep(u, v) = \chi(u) \cap \chi(v)$; *the* eliminator *of $u$ and $v$ is defined as* $elim(u, v) = \chi(u) - sep(u, v)$.

DEFINITION **2.3 (tree-width, hyper-width, induced-width)** *The* tree-width *of a tree-decomposition is* $tw = \max_{v \in V} |\chi(v)| - 1$, *its* hyper-width *is* $hw = \max_{v \in V} |\psi(v)|$, *and its* maximum separator size *is* $s = \max_{(u,v) \in E} |sep(u,v)|$. *The tree-width of a graph is the minimum tree-width over all its tree-decompositions and is identical to the graph's induced-width.*

## 2.2 CLUSTER-TREE ELIMINATION

Algorithm Cluster-Tree Elimination (CTE) provides a unifying space concious description of join-tree clustering algorithms. It is a message-passing scheme that runs on the tree-decomposition, well-known for solving a wide range of automated reasoning problems. We will briefly describe its partition-based mini-clustering approximation that forms the basis for our heuristic generation scheme.

CTE provided in Figure 1 computes a solution to the singleton functions $z_i$ in a Bayesian network. It works by computing *messages* that are sent along edges in the tree. Message $m_{(u,v)}$ sent from vertex $u$ to vertex $v$, can be computed as soon as all incoming messages to $u$ other than $m_{(v,u)}$ have been received. As leaves compute their messages, their adjacent vertices also qualify and computation goes on until all messages have been computed. The set of functions associated with a vertex $u$ augmented with the set of incoming messages is called a *cluster*, $cluster(u) = \psi(u) \cup_{(w,u) \in T} m_{(w,u)}$. A message $m_{(u,v)}$ is computed as the product of all functions in $cluster(u)$ excluding $m_{(v,u)}$ and the subsequent elimination of variables in the eliminator of $u$ and $v$. Formally, $m_{(u,v)} = \max_{elim(u,v)} (\prod_{g \in cluster(u), g \neq m_{(v,u)}} g)$. The computation is done by enumeration, recording only the output message. The algorithm terminates when all messages are computed.

The functions $z(X_i)$ can be computed in any cluster that contains $X_i$ by eliminating all variables other than $X_i$.

It was shown that [Dechter *et al.*2001] the complexity of CTE is time $O(r \cdot (hw + dg) \cdot d^{tw+1})$ and space $O(r \cdot d^s)$, where $r$ is the number of vertices in the tree-decomposition, $hw$ is the hyper-width, $dg$ is the maximum degree (i.e., number of adjacent vertices) in the tree, $tw$ is the tree-width, $d$ is the largest domain size and $s$ is the maximum separator size. This assumes that step 2 is computed by enumeration.

There is a variety of ways in which a tree-decomposition can be obtained. We will choose a particular one called bucket-tree decomposition, inspired by viewing the bucket-elimination algorithm as message passing along a tree [Dechter *et al.*2001]. Since bucket-tree is a special case of a cluster-tree, we define the CTE algorithm applied to a bucket-tree to be called Bucket-Tree Elimination (BTE). BTE has time and space complexity $O(r \cdot d^{tw+1})$.

## 2.3 MINI-CLUSTER-TREE ELIMINATION

The main drawback of CTE and any variant of join-tree algorithms is that they are time and space exponential in the tree-width ($tw$) and separator ($s$) size, respectively [Dechter *et al.*2001, Mateescu *et al.*2002], which are often very large. In order to overcome this problem, partition-based algorithms were introduced. Instead of combining all the functions in a cluster, when computing a message, we first partition the functions in the cluster into a set of mini-clusters such that each mini-cluster is bounded by a fixed number of variables ($i$-bound), and then process them separately. The algorithm, called Mini-Cluster-Tree Elimination (MCTE) approximates CTE and it computes upper bounds on values computed by CTE.

In the Mini-Cluster-Tree Elimination the message $M_{(u,v)}$ that node $u$ sends to node $v$ is a set of functions computed as follows. The functions in $cluster(u) - M_{(v,u)}$ are partitioned into $\mathbf{P} = \mathbf{P_1}, \cdots, \mathbf{P_k}$, where $|\text{scope}(\mathbf{P_j})| \leq i$, for a given $i$. The message $M_{(u,v)}$ is defined as $M_{(u,v)} = \{\max_{elim(u,v)} \prod_{g \in \mathbf{P_j}} g | \mathbf{P_j} \in \mathbf{P}\}$. Algorithm MCTE applied to the bucket-tree is called Mini-Bucket-Tree elimination (MBTE) [Dechter *et al.*2001].

Since the scope size of each mini-cluster is bounded by $i$, the time and space complexity of MCTE (MBTE) is expo-

---

**Procedure BBBT($\mathcal{T}$,$i$,$s$,$L$)**
**Input:** Bucket-tree $\mathcal{T}$, parameter $i$, set of instantiated variables $S = s$, lower bound $L$.
**Output:** MPE probability conditioned on $s$.
1. If $S = X$, return the probability of the current complete assignment.
2. **Run** MBTE($i$); Let $\{mz_j\}$ be the set of heuristic values computed by MBTE($i$) for each $X_j \in X - S$.
3. **Prune** domains of uninstantiated variables, by removing values $x \in D(X_l)$ for which $mz_l(x) \leq L$.
4. **Backtrack:** If $D(X_l) = \emptyset$ for some variable $X_l$, return 0.
5. **Otherwise** let $X_j$ be the uninstantiated variable with the smallest domain: $X_j = argmin_{X_k \in X-S}|D(X_k)|$.
6. **Repeat** while $D(X_j) \neq \emptyset$
    $i$. Let $x_k$ be the value of $X_j$ with the largest heuristic estimate: $x_k = argmax_{x_j \in D(X)} mz_j(x_j)$.
    $ii$. Set $D(X) = D(X) - x_k$.
    $iii$. Compute $mpe = BBBT(\mathcal{T}, i, s \cup \{X_j = x_k\}, L)$.
    $iv$. Set $L = max(L, mpe)$.
    $v$. Prune $D(X_j)$ by $L$.
7. **Return** $L$.

---

Figure 2: Branch-and-Bound with MBTE (BBBT).

nential in $i$. However, because of the partitioning, the functions $z_j$ cannot be computed exactly any more. Instead, the output functions of MCTE (MBTE), called $mz_j$, are upper bounds on the exact functions $z_j$ ([Dechter et al. 2001]).

Clearly, increasing $i$ is likely to provide better upper bounds at a higher cost. Therefore, MCTE($i$) allows trading upper bound accuracy for time and space complexity.

## 3 PARTITION-BASED BnB

This section focuses on the two systematic algorithms we used. Both use partition based mini-bucket heuristics.

### 3.1 BnB WITH DYNAMIC HEURISTICS (BBBT)

Since MBTE($i$) computes upper bounds for each singleton-variable assignment simultaneously, when incorporated within a depth-first Branch-and-Bound algorithm, MBTE($i$) can facilitate domain pruning and dynamic variable ordering.

Such a Branch-and-Bound algorithm, called BBBT($i$), for solving the MPE problem is given in Figure 2. Initially it is called with BBBT($< \mathcal{T}, \chi, \psi >, i, \emptyset, 0$). At all times it maintains a lower bound $L$ which corresponds to the probability of the best assignment found so far. At each step, it executes MBTE($i$) which computes the singleton assignment costs $mz_i$ for each uninstantiated variable $X_i$ (step 2), and then uses these costs to prune the domains of uninstantiated variables by comparing $L$ against the heuristic estimate of each value (step 3). If the cost of the value is not more than $L$, it can be pruned because it is an upper bound. If as a result a domain of a variable becomes empty, then the current partial assignment is guaranteed not to lead to a better assignment and the algorithm can backtrack (step 4). Otherwise, BBBT expands the current assignment picking a variable $X_j$ with the smallest domain (variable ordering in step 5) and recursively solves a set of subproblems, one for each value of $X_j$, in decreasing order of heuristic estimates of its values (value ordering in step 6). If during the solution of the subproblem a better new assignment is found, the lower bound $L$ can be updated (step 6$iv$).

Thus, at each node in the search space, BBBT($i$) first executes MBTE($i$), then prunes domains of all un-instantiated variables, and then recursively solves a set of subproblems. BBBT performs a look-ahead computation that is similar (but not identical) to $i$-consistency at each search node.

### 3.2 BnB WITH STATIC MINI-BUCKETS (BBMB)

As described in the introduction, the strength of BBMB($i$) was well established in several empirical studies [Kask and Dechter 2001]. We describe the main differences between BBBT and BBMB:

- BBMB($i$) uses as a pre-processing step the Mini-Bucket-Elimination, which compiles a set of functions that can be used to assemble efficiently heuristic estimates during search. The main overhead is therefore the pre-processing step which is exponential in the $i$-bound but does not depend on the number of search nodes. BBBT($i$) on the other hand computes the heuristic estimates solely during search using MBTE($i$). Consequently its overhead is exponential in the $i$-bound multiplied by the number of nodes visited.

- Because of the pre-computation of heuristics, BBMB is limited to static variable ordering, while BBBT uses a dynamic variable ordering.

- Finally, since at each step, BBBT computes heuristic estimates for all un-instantiated variables, it can prune their domains, which provides a form of look-ahead. BBMB on the other hand generates a heuristic estimate only for the next variable in the static ordering and prunes only its domain.

## 4 NON-SYSTEMATIC ALGORITHMS

This section focuses on two different types of incomplete algorithms: stochastic local search and iterative belief propagation.

### 4.1 LOCAL SEARCH

Local search is a general optimization technique which can be used alone or as a method for improving solutions found by other approximation scheme. Unlike the Branch-and-Bound algorithms, these methods do not guarantee an optimal solution. [Park 2002] showed that an MPE problem can be converted to a weighted CNF expression whose MAX-SAT solution immediately produces the solution of the corresponding MPE problem. Subsequently, local search algorithms initially developed for the weighted MAX-SAT domain can be used for approximating the MPE problem in Bayesian networks. We continue the investigation of Guided Local Search (GLS) and Discrete Lagrangian Multipliers (DLM) algorithms, as well as a previous approach (SLS) proposed in [Kask and Dechter 1999].

The method of **Discrete Lagrangian Multipliers** [Wah and Shang 1997] is based on an extension of constraint optimization using Lagrange multipliers for continuous variables. In the weighted MAX-SAT domain, the clauses are the constraints, and the sum of the unsatisfied clauses is the cost function. In addition to the weight $w_C$, a Lagrangian multiplier $\lambda_C$ is associated with each clause. The cost function for DLM is of the form:

$$\sum_C w_C + \sum_C \lambda_C$$

where $C$ ranges over the unsatisfied clauses. Every time a local maxima is encountered, the $\lambda$s corresponding to the unsatisfied clauses are incremented by a adding a constant. **Guided Local Search** [Mills and Tsang 2000] is a heuristically developed method for solving combinatorial optimization problems. It has been shown to be extremely efficient at solving general weighted MAX-SAT problems. Like DLM, GLS associates an additional weight with each clause C ($\lambda_C$). The cost function in this case is essentially $\sum_C \lambda_C$, where $C$ ranges over the unsatisfied clauses. Every time a local maxima is reached, the $\lambda$s of the unsatisfied clauses with maximum utility are increased by adding a constant, where the utility of a clause C is given by $w_C/(1 + \lambda_C)$. Unlike DLM, which increments all the weights of the unsatisfied clauses, GLS modifies only a few of them.
**Stochastic Local Search** is a local search algorithm that at each step performs either a hill climbing or a stochastic variable change. Periodically, the search is restarted in order to escape local maxima. It was shown to be superior to simulated annealing and some pure greedy search algorithms.

In [Park 2002] it was shown that, among these three algorithms, GLS provided the best overall performance on a variety of problem classes, both random and real-world benchmarks.

### 4.2 ITERATIVE JOIN-GRAPH PROPAGATION

The **Iterative Join Graph Propagation** (IJGP) [Dechter et al. 2002] algorithm belongs to the class of generalized belief propagation methods, recently proposed to generalize Pearl's belief propagation algorithm [Pearl 1988] using analogy with algorithms in statistical physics. This class of algorithms, developed initially for belief updating, is an iterative approximation method that applies the message passing algorithm of join-tree clustering to join-graphs, iteratively. It uses a parameter $i$ that bounds the complexity and makes the algorithm anytime. Here, we adapted the IJGP($i$) algorithm for solving the MPE problem by replacing the sum-product messages with max-product message propagation.

## 5 EXPERIMENTAL RESULTS

We tested the performance of our scheme for solving the MPE task on several types of belief networks - random uniform and Noisy-OR Bayesian networks, NxN grids, coding networks, CPCS networks and 9 real world networks obtained from the Bayesian Network Repository[1]. On each problem instance we ran BBBT($i$) and BBMB($i$) with various $i$-bounds, as heuristics generators, as well as the local search algorithms discussed earlier. We also ran the Iterative Join Graph Propagation algorithm (IJGP) on some of these problems.

We treat all algorithms as approximation schemes. Algorithms BBBT and BBMB have any-time behavior and, if allowed to run until completion, will solve the problem exactly. However, in practice, both algorithms may be terminated at a time bound and may return sub-optimal solutions. On the other hand, neither the local search techniques, nor the belief propagation algorithms guarantee an optimal solution, even if given enough time.

To measure performance we used the accuracy ratio $opt = P_{alg} / P_{MPE}$ between the value of the solution found by the test algorithm ($P_{alg}$) and the value of the optimal solution ($P_{MPE}$), whenever $P_{MPE}$ was available. We only report results for the range $opt \geq 0.95$. We also recorded the average running time for all algorithms, as well as the average number of search tree nodes visited by the Branch-and-Bound algorithms. When the size and difficulty of the problem did not allow an exact computation, we compared the quality of the solutions produced by the respective algorithms in the given time bound. For each problem class we chose a number of evidence variables, randomly and fixed their values.

---

[1] www.cs.huji.ac.il/labs/compbio/Repository

| K | BBBT BBMB IJGP i=2 %[time]{nodes} | BBBT BBMB IJGP i=4 %[time]{nodes} | BBBT BBMB IJGP i=6 %[time]{nodes} | BBBT BBMB IJGP i=8 %[time]{nodes} | BBBT BBMB IJGP i=10 %[time]{nodes} | GLS %[time] | DLM %[time] | SLS %[time] |
|---|---|---|---|---|---|---|---|---|
| 2 | 90[6.30]{3.9K} 71[2.19]{1.6M} 62[0.04] | 100[1.19]{781} 92[0.17]{0.1M} 66[0.06] | 100[0.65]{366} 92[0.02]{10K} 66[0.13] | **100[0.44]{212}** 86[0.01]{3K} 71[0.32] | 100[0.43]{161} 91[0.01]{1.2K} 67[0.87] | 100[1.05] | 0[30.01] | 0[30.01] |
| 3 | 28[46.6]{19K} 5[43.1]{16M} 34[0.07] | 65[27.5]{5.5K} 78[24.4]{8.2M} 37[0.18] | 86[15.4]{1.1K} **90[3.20]{0.8M}** 36[0.94] | 86[19.3]{453} 89[1.23]{0.3M} 43[5.38] | 80[27.5]{213} 83[0.58]{52.5K} 44[32.5] | 39[44.02] | 0[60.01] | 0[60.01] |
| 4 | 24[95.5]{63K} 3[89.4]{47M} 17[0.14] | 46[74.7]{13.4K} 42[85.5]{37M} 14[0.47] | 65[54.1]{1.6K} 89[25.4]{8M} 14[4.33] | 67[65.7]{443} **90[5.44]{1.5M}** 17[43.3] | 37[151.2]{74} 99[4.82]{0.3M} 20[468.5] | 5[114.9] | 0[120.01] | 0[120.01] |

Table 1: Average accuracy and time. Random Bayesian (N=100, C=90, P=2). w*=17, 10 evidence, 100 samples.

## 5.1 RANDOM BAYESIAN NETWORKS AND NOISY-OR NETWORKS

The random Bayesian networks were generated using parameters (N, K, C, P), where N is the number of variables, K is their domain size, C is the number of conditional probability tables (CPTs) and P is the number of parents in each CPT. The structure of the network is created by randomly picking C variables out of N and, for each, randomly selecting P parents from their preceding variables, relative to some ordering. For random uniform Bayesian networks, each probability table is generated uniformly randomly. For Noisy-OR networks, each probability table represents an OR-function with a given *noise* and *leak* probabilities: $P(X = 0|Y_1, \ldots, Y_p) = P_{leak} \times \prod_{Y_i=1} P_{noise}$

Tables 1 and 2 present experiments with random uniform Bayesian networks and Noisy-OR networks, respectively. In each table, parameters N, C and P are fixed, while K, controlling the domain size of the network's variables, is changing. For each value of K, we generate 100 instances. We gave each algorithm a time limit of 30, 60 and 120 seconds, depending on the value of the domain size. Each test case had 10 randomly selected evidence variables. We have highlighted the best performance point in each row.

For example, Table 1 reports the results with random problems having N=100, C=90, P=2. Each horizontal block corresponds to a different value of K. The columns show results for BBBT/BBMB/IJGP at various levels of $i$, as well as for GLS, DLM and SLS. Looking at the first line in Table 1 we see that in the accuracy range $opt \geq 0.95$ and for the smallest domain size (K = 2) BBBT with $i$=2 solved 90% of the instances using 6.30 seconds on average and exploring 3.9K nodes, while BBMB with $i$=2 only solved 71% of the instances using 2.19 seconds on average and exploring a much larger search space (1.6M nodes). GLS significantly outperformed the other local search methods, as also observed in [Park 2002] and solved all instances using 1.05 seconds on average. However, as BBBT($i$)'s bound increases, it is better than GLS. As the domain size increases, the problem instances become harder. The overall performance of local search algorithms, especially GLS's performance, deteriorates quite rapidly.

When comparing BBBT($i$) to BBMB($i$) we notice that at larger domain sizes (K ∈ {3, 4}) the superiority of BBBT($i$) is more pronounced for small $i$-bounds ($i$=2,4), both in terms of the quality of the solution and search space explored. This may be significant, because small $i$-bounds require restricted space.

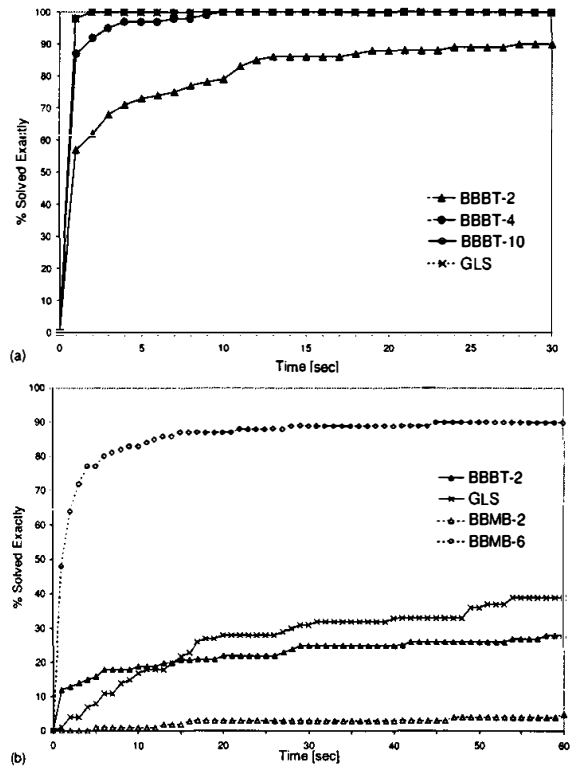

Figure 3: Random Bayesian (N=100, C=90, P=2) (a) K=2 (b) K=3. 10 evidence, 100 samples.

In Figure 3 we provide an alternative view of the performance of BBBT($i$)/BBMB($i$) against GLS as anytime algorithms. Let $F_{alg}(t)$ be the fraction of problems solved completely by the test algorithm $alg$ by time $t$. Each graph in Figure 3 plots $F_{BBBT(i)}(t)$, $F_{BBMB(i)}(t)$ for some selected values of $i$, as well as $F_{GLS}(t)$. Two different values of the domain size are discussed, K=2 and K=3, respectively. Figure 3 shows the distributions of $F_{BBBT(i)}(t)$, $F_{BBMB(i)}(t)$ and $F_{GLS}(t)$ for the random uniform Bayesian networks when N=100, C=90, P=2 (corresponding to the first two rows in Table 1).

| K | BBBT BBMB IJGP i=2 %[time] | BBBT BBMB IJGP i=4 %[time] | BBBT BBMB IJGP i=6 %[time] | BBBT BBMB IJGP i=8 %[time] | BBBT BBMB IJGP i=10 %[time] | GLS %[time] | DLM %[time] | SLS %[time] |
|---|---|---|---|---|---|---|---|---|
| 2 | 84[7.34] 61[3.49] 62[0.04] | 98[2.48] 91[0.30] 66[0.06] | 100[0.88] 89[0.05] 66[0.13] | 100[0.66] 88[0.02] 71[0.31] | 100[0.59] 88[0.02] 67[0.86] | 100[1.25] | 0[30.02] | 0[30.02] |
| 3 | 36[42.2] 8[47.5] 34[0.04] | 78[19.1] 77[18.4] 37[0.10] | 95[9.64] 95[1.81] 36[0.49] | 94[10.7] 86[0.71] 43[2.86] | 93[16.8] 84[0.33] 44[17.0] | 49[38.7] | 0[60.02] | 0[60.01] |
| 4 | 24[97.7] 2[114.4] 17[0.06] | 40[80.3] 39[92.3] 14[0.23] | 61[62.4] 84[33.2] 14[2.12] | 58[82.0] 90[7.39] 17[21.9] | 30[269] 99[7.95] 20[226.8] | 5[115.03] | 0[120.01] | 0[120.01] |

Table 2: Average accuracy and time. Random Noisy-OR (N=100, C=90, P=2). $P_{noise} = 0.2, P_{leak} = 0.01$. w*=17, 10 evidence, 100 samples.

| K | BBBT/GLS BBMB/GLS i=2 # best | BBBT/GLS BBMB/GLS i=3 # best | BBBT/GLS BBMB/GLS i=4 # best | BBBT/GLS BBMB/GLS i=5 # best | BBBT/GLS BBMB/GLS i=6 # best |
|---|---|---|---|---|---|
| 2 | 0/29 0/24 | 0/25 0/19 | 0/23 0/19 | 0/21 0/5 | 0/20 0/5 |
| 3 | 4/26 1/29 | 5/25 2/28 | 5/25 2/28 | 9/21 2/28 | 10/20 4/26 |
| 5 | 28/2 5/25 | 28/2 5/25 | 30/0 7/23 | 30/0 12/18 | 30/0 23/7 |
| 7 | 25/5 18/12 | 22/8 15/15 | 24/6 17/13 | 19/11 20/10 | 21/9 25/5 |

Table 3: # wins given fixed time bound. Random Networks (N=100,C=90,P=3). w*=30, 10 evidence, 30 samples.

Clearly, if $F_{alg^i}(t) > F_{alg^j}(t)$, then $F_{alg^i}(t)$ completely dominates $F_{alg^j}(t)$. For example, in Figure 3(a), GLS is highly competitive with BBBT(10) and both significantly outperform BBBT(i)/BBMB(i) for smaller i-bounds. In contrast, Figure 3(b) shows how the best local search method deteriorates as the domain size increases.

We also experimented with a much harder set of random Bayesian networks. The dataset consisted of random networks with parameters N=100, C=90, P=3. In this case, the induced width of the problem instances was around 30, thus it was not possible to compute exact solutions. We studied four domain sizes K∈{2, 3, 5, 7}. For each value of K, we generate 30 problem instances. Each algorithm was allowed a time limit of 30, 60, 120 and 180 seconds, depending on the domain size. We found that the solutions generated by DLM and SLS were several orders of magnitude smaller than those found by GLS, BBBT and BBMB. Hence, we only report the latter three algorithms.

Table 3 compares the frequency that the solution was the best for each of the three algorithms (ties are removed). We notice again that GLS excelled at finding the best solution for smaller domain sizes, in particular for K=2 and 3. On the other hand, at larger domain sizes (K∈{5,7}), the power of BBBT(i) is more pronounced for smaller i-bounds, whereas BBMB(i) is more efficient at larger i-bounds. Figure 4 shows, pictorially, the quality of the solutions produced by GLS against the ones produced by BBBT(i)/BBMB(i). For each plot, corresponding to a different value of the domain size K, the X axis represents the negative log probability of the solutions found by GLS and the Y axis represents the negative log probability of the solutions found by BBBT(i)/BBMB(i). The superiority of BnB-based methods for larger domain sizes is significant since these are algorithms that can prove optimality when given enough time, unlike local search methods. Not if the superiority of GLS for K∈{2,3} may be a function of the time bound for these hard instances.

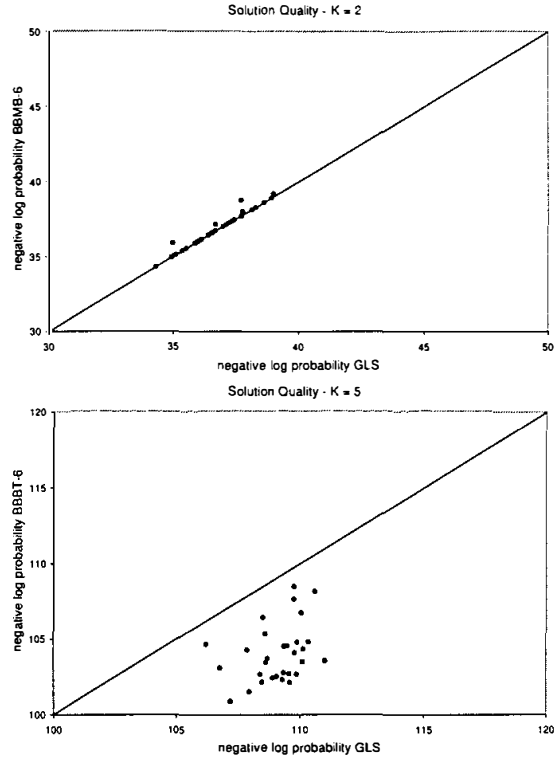

Figure 4: Solution quality at fixed time bound. Random Networks (N=100, C=90, K, P=3). w*=30, 10 evidence, 100 samples. (a)K=2 (b)K=5

| K | BBBT/BBMB i=2 # wins # nodes | BBBT/BBMB i=3 # wins # nodes | BBBT/BBMB i=4 # wins # nodes | BBBT/BBMB i=5 # wins # nodes |
|---|---|---|---|---|
| 2 | 20/10 15.3K/13.8M | 12/18 14.5K/16.2M | 18/12 12.3K/15.9M | 18/12 9.4K/11.8M |
| 3 | 27/3 19.9K/16.3M | 26/4 13.8K/16.8M | 29/1 12.3K/15.9M | 28/2 4.8K/14.2 |
| 5 | 29/1 18.3K/10.5M | 30/0 9.1K/13.8M | 30/0 3.5K/13.2M | 27/3 0.9K/12.6M |
| 7 | 24/6 7.7K/8.3M | 26/4 3.4K/10.6M | 14/16 114/10.9M | 10/20 8/9.6M |

Table 4: BBBT vs. BBMB. Random networks (N=100, C=90, P=3). 10 evidence, 30 samples, 30 seconds.

Table 4 shows comparatively the performance of BBBT as compared to BBMB. Each entry in the table shows the number of times BBBT produced a better solution than BBMB (# wins) as well as the average number of search tree nodes visited by both algorithms.

## 5.2 GRID NETWORKS

In grid networks, N is a square number and each CPT is generated uniformly randomly or as a Noisy-OR function. We experimented with a synthetic set of 10-by-10 grid networks. We report results on three different domain sizes.

| K | BBBT/ BBMB i=2 %[time] | BBBT/ BBMB i=4 %[time] | BBBT/ BBMB i=6 %[time] | BBBT/ BBMB i=8 %[time] | BBBT/ BBMB i=10 %[time] | GLS %[time] | DLM %[time] | SLS %[time] |
|---|---|---|---|---|---|---|---|---|
| 2 | 51[17.7] 1[29.9] | 99[2.62] 13[23.7] | 100[0.66] 93[2.16] | 100[0.48] 92[0.08] | 100[0.42] 95[0.02] | 100[1.54] | 0[30.01] | 0[30.01] |
| 3 | 3[58.7] 0[60.01] | 28[47.4] 1[58.9] | 80[19.5] 25[50.9] | 93[14.8] 89[8.63] | 94[23.2] 92[0.73] | 4[58.7] | 0[60.01] | 0[60.01] |
| 4 | 1[118.8] 0[120] | 12[108.3] 0[120] | 46[78.4] 6[113.4] | 61[88.5] 72[46.4] | 33[136] 85[9.91] | 0[120] | 0[120] | 0[120] |

Table 5: Average accuracy and time. Random Grid Networks (N=100). w*=15, 10 evidence, 100 samples.

| $\sigma$ | BBBT BBMB IJGP i=2 BER[time] | BBBT BBMB IJGP i=4 BER[time] | BBBT BBMB IJGP i=6 BER[time] | BBBT BBMB IJGP i=8 BER[time] | BBBT BBMB IJGP i=10 BER[time] | IBP GLS SLS BER[time] |
|---|---|---|---|---|---|---|
| 0.32 | 0.0056[3.18] 0.0034[0.07] 0.0034[0.16] | 0.0104[2.87] 0.0034[0.08] 0.0034[0.18] | 0.0072[1.75] 0.0034[0.03] 0.0034[0.33] | 0.0034[0.72] 0.0034[0.01] 0.0034[0.92] | 0.0034[0.59] 0.0034[0.02] 0.0034[3.02] | 0.0034[0.01] 0.2344[60.01] 0.4980[60.01] |
| 0.40 | 0.0642[19.4] 0.0114[0.63] 0.0114[0.16] | 0.0400[12.8] 0.0114[0.53] 0.0138[0.18] | 0.0262[6.96] 0.0114[0.12] 0.0118[0.33] | 0.0148[4.52] 0.0114[0.05] 0.0116[0.91] | 0.0190[4.34] 0.0114[0.04] 0.0120[3.02] | 0.0108[0.01] 0.2084[60.01] 0.5128[60.01] |
| 0.52 | 0.1920[48.1] 0.0948[1.35] 0.1224[0.08] | 0.1790[42.0] 0.0948[1.47] 0.1242[0.09] | 0.1384[31.3] 0.0948[0.36] 0.1256[0.16] | 0.1144[21.4] 0.0948[0.11] 0.1236[0.47] | 0.1144[19.7] 0.0948[0.05] 0.1132[1.54] | 0.0894[0.01] 0.2462[60.02] 0.5128[60.01] |

Table 6: Average BER and time. Random Coding Networks (N=200, P=4). w*=22, 60 seconds, 100 samples.

For each value of K, we generate 100 problem instances. Each algorithm was allowed a time limit of 30, 60 and 120 seconds, depending on the domain size. Table 5 shows the average accuracy and running time for each algorithm.

### 5.3 RANDOM CODING NETWORKS

Our random coding networks fall within the class of *linear block codes*. They can be represented as four-layer belief networks having K nodes in each layer. The decoding algorithm takes the coding network as input and the observed channel output and computes the MPE assignment. The performance of the decoding algorithm is usually measured by the Bit Error Rate (BER), which is simply the observed fraction of information bit errors.

We tested random coding networks with K=50 input bits and various levels of channel noise $\sigma$. For each value of $\sigma$ we generate 100 problem instances. Each algorithm was allowed a time limit of 60 seconds. Table 6 reports the average Bit Error Rate, as well as the average running time of the algorithms. We see that BBBT/BBMB outperformed considerably GLS. On the other hand, only BBMB is competitive to IBP, which is the best performing algorithm for coding networks.

### 5.4 REAL WORLD NETWORKS

Our realistic domain contained 9 Bayesian networks from the Bayesian Network Repository, as well as 4 CPCS networks derived from the Computer-Based Care Simulation system. For each network, we ran 20 test cases. Each test case had 10 randomly selected evidence variables, ensuring that the probability of evidence was positive. Each algorithm was allowed a 30 second time limit.

Table 7 summarizes the results. For each network, we list the number of variables, the average and maximum domain size for its variables, as well as the induced width. We also provide the percentage of exactly solved problem instances and the average running time for each algorithm.

In terms of accuracy, we notice a significant dominance of the systematic algorithms over the local search methods, especially for networks with large domains (e.g. Barley, Mildew, Diabetes, Munin). For networks with relatively small domain sizes (e.g. Pigs, Water, CPCS networks) the non-systematic algorithms, in particular GLS, solved almost as many problem instances as the Branch-and-Bound algorithms. Nevertheless, the running time of BBBT/BBMB was much better in this case, because GLS had to run until exceeding the time limit, even though it might have found the optimal solution within the first few iterations. BBBT/BBMB on the other hand terminated, hence proving optimality.

We also used for comparison the IJGP algorithm, set up for 30 iterations. In terms of average accuracy, we notice the stable performance of the algorithm in almost all test cases. For networks with large domain sizes, IJGP($i$) significantly dominated the local search algorithms and in some cases it even outperformed the BBBT($i$)/BBMB($i$) algorithms (e.g. Barley, Mildew, Munin).

## 6 CONCLUSION

The paper investigates the performance of two Branch-and-Bound search algorithms (BBBT/BBMB) against a number of state-of-the-art stochastic local search (SLS) algorithms for the problem of solving the MPE task in Bayesian networks. Both BBBT and BBMB use the idea of partioning-based approximation of inference for heuristic computation, but in different ways: while BBMB uses a static pre-computed heuristic function, BBBT computes it dynamically at each step. We observed over a wide range of problem classes, both random and real-world benchmarks, that BBBT/BBMB are often superior to SLS, except in cases when the domain size is small, in which case they are competitive. This is in stark contrast with the performance of systematic vs. non-systematic on CSP/SAT problems, where SLS algorithms often significantly outperform complete methods. An additional advantage of BBBT/BBMB is that as complete algorithms they can prove optimality if given enough time, unlike SLS.

When designing algorithms to solve an NP-hard task, one cannot hope to develop a single algorithm that would be superior across all problem classes. Our experiments show that BBBT/BBMB, when viewed as a collection of algorithms parametrized by $i$, show robust performance over a wide range of MPE problem classes, because for each problem instance there is a value of $i$, such that the performance of BBBT($i$)/BBMB($i$) dominates that of SLS.

| Network | # vars | avg. dom. | max dom. | w* | BBBT/ BBMB/ IJGP i=2 %[time] | BBBT/ BBMB/ IJGP i=3 %[time] | BBBT/ BBMB/ IJGP i=4 %[time] | BBBT/ BBMB/ IJGP i=5 %[time] | BBBT/ BBMB/ IJGP i=6 %[time] | BBBT/ BBMB/ IJGP i=7 %[time] | BBBT/ BBMB/ IJGP i=8 %[time] | BBBT/ BBMB/ IJGP i=10 %[time] | GLS % [time] | DLM % [time] | SLS % [time] |
|---|---|---|---|---|---|---|---|---|---|---|---|---|---|---|---|
| Barley | 48 | 8 | 67 | 8 | -<br>-<br>67[0.99] | -<br>-<br>67[1.11] | 90[6.33]<br>25[12.8]<br>63[1.49] | 100[4.28]<br>40[2.32]<br>70[5.32] | 100[3.29]<br>65[0.43]<br>80[17.9] | 100[2.81]<br>90[0.85]<br>- | 100[2.91]<br>100[2.41]<br>- | -<br>-<br>- | 0<br>[30.01] | 0<br>[30.01] | 0<br>[30.01] |
| Diabetes* | 413 | 11 | 21 | 5 | 0[120]<br>0[120]<br>3[8.60] | 0[123]<br>0[120]<br>3[11.2] | 0[127]<br>5[114]<br>43[86.0] | 90[21.1]<br>100[2.01]<br>97[311.1] | -<br>-<br>100[384.6] | -<br>-<br>- | -<br>-<br>- | -<br>-<br>- | 0<br>[120.01] | 0<br>[120.01] | 0<br>[120.01] |
| Mildew | 35 | 17 | 100 | 4 | 100[0.28]<br>30[10.5]<br>90[3.59] | 100[0.17]<br>65[7.5]<br>87[3.68] | 100[0.56]<br>95[0.18]<br>97[33.3] | -<br>-<br>100[53.2] | -<br>-<br>- | -<br>-<br>- | -<br>-<br>- | -<br>-<br>- | 15<br>[30.02] | 0<br>[30.02] | 90<br>[30.02] |
| Munin1 | 189 | 5 | 21 | 11 | 90[6.13]<br>0[30]<br>90[0.45] | -<br>-<br>90[0.49] | 100[6.48]<br>5[27.2]<br>97[1.10] | -<br>-<br>93[4.28] | 40[23.8]<br>20[24.1]<br>93[14.5] | -<br>-<br>97[70.2] | 75[13.4]<br>70[6.77]<br>100[191.9] | 80[43.1]<br>100[9.03]<br>- | 10<br>[30.02] | 0<br>[30.02] | 0<br>[30.02] |
| Munin2 | 1003 | 5 | 21 | 7 | 95[1.65]<br>95[30.3]<br>95[2.44] | 95[1.73]<br>95[31.7]<br>95[2.94] | 95[1.65]<br>95[30.5]<br>95[5.17] | 95[1.99]<br>95[31.8]<br>100[20.3] | 95[2.32]<br>95[31.3]<br>95[64.9] | 95[2.48]<br>100[30.5]<br>- | 100[1.97]<br>100[1.84]<br>- | -<br>-<br>- | 0<br>[30.01] | 0<br>[30.01] | 0<br>[30.01] |
| Munin3 | 1044 | 5 | 21 | 7 | 0[30.8]<br>0[30.2]<br>80[1.47] | 0[30.9]<br>0[31]<br>95[1.72] | 0[31.3]<br>0[32.3]<br>85[3.10] | 5[31.7]<br>5[29.9]<br>85[10.8] | 0[40.9]<br>0[32.7]<br>90[38.9] | 90[4.72]<br>95[2.14]<br>- | 100[2.2]<br>100[1.01]<br>- | -<br>-<br>- | 0<br>[30.02] | 0<br>[30.02] | 0<br>[30.02] |
| Munin4 | 1041 | 5 | 21 | 8 | 0[31]<br>0[30.2]<br>85[1.52] | 0[31]<br>0[31.4]<br>75[1.66] | 0[31.9]<br>0[31.6]<br>90[4.15] | 0[37.7]<br>0[32]<br>95[15.6] | 0[44.5]<br>0[30.3]<br>95[43.6] | 0[58.8]<br>30[22.1]<br>- | 0[170.4]<br>85[3.4]<br>- | -<br>-<br>- | 0<br>[30.02] | 0<br>[30.02] | 0<br>[30.02] |
| Pigs | 441 | 3 | 3 | 12 | 90[15.2]<br>0[30.01]<br>80[0.31] | -<br>-<br>73[0.37] | 100[3.73]<br>60[4.85]<br>77[0.53] | -<br>-<br>83[0.86] | 100[2.36]<br>80[0.02]<br>80[1.43] | -<br>-<br>80[2.49] | 100[0.58]<br>95[0.04]<br>83[6.27] | 100[0.56]<br>95[0.12]<br>93[27.3] | 10<br>[30.02] | 0<br>[30.02] | 0<br>[30.02] |
| Water | 32 | 3 | 4 | 11 | 100[0.01]<br>55[4.51]<br>97[0.09] | -<br>-<br>97[0.09] | 100[0.02]<br>60[4.5]<br>97[0.10] | -<br>-<br>97[0.14] | 100[0.03]<br>75[0.01]<br>100[0.26] | -<br>-<br>100[0.45] | 100[0.04]<br>100[0.02]<br>100[1.12] | 100[0.09]<br>100[0.06]<br>100[5.94] | 100<br>[30.02] | 75<br>[30.02] | 100<br>[30.02] |
| CPCS54 | 54 | 2 | 2 | 15 | 100[0.35]<br>35[0.02]<br>67[0.06] | -<br>-<br>77[0.06] | 100[0.18]<br>60[0.01]<br>67[0.06] | -<br>-<br>70[0.07] | 100[0.11]<br>50[0.01]<br>63[0.09] | -<br>-<br>70[0.11] | 100[0.09]<br>55[0.004]<br>63[0.16] | 100[0.06]<br>60[0.003]<br>73[0.38] | 100<br>[30.02] | 0<br>[30.02] | 100<br>[30.02] |
| CPCS179 | 179 | 2 | 4 | 8 | 100[1.69]<br>80[0.02]<br>100[2.50] | -<br>-<br>100[2.52] | 100[1.01]<br>80[0.02]<br>100[2.99] | -<br>-<br>100[3.37] | 100[0.05]<br>100[0.02]<br>100[6.49] | -<br>-<br>100[8.63] | 100[0.11]<br>100[0.07]<br>100[36.9] | -<br>-<br>- | 100<br>[30.02] | 30<br>[30.02] | 30<br>[30.02] |
| CPCS360b | 360 | 2 | 2 | 20 | 100[0.17]<br>100[0.04]<br>100[10.6] | -<br>-<br>100[10.4] | 100[0.27]<br>100[0.03]<br>100[10.5] | -<br>-<br>100[10.1] | 100[0.21]<br>100[0.03]<br>100[9.82] | -<br>-<br>100[8.19] | 100[0.19]<br>100[0.03]<br>100[8.59] | 100[0.32]<br>100[0.04]<br>100[12.5] | 100<br>[30.02] | 100<br>[30.02] | 100<br>[30.02] |
| CPCS422b* | 422 | 2 | 2 | 23 | 65[52.6]<br>100[0.5]<br>83[88.0] | -<br>-<br>83[86.8] | 70[48.7]<br>100[0.49]<br>87[86.4] | -<br>-<br>90[84.3] | 70[47.2]<br>100[0.49]<br>83[85.3] | -<br>-<br>87[77.7] | 90[21.5]<br>100[0.47]<br>87[77.1] | 95[12.9]<br>100[0.47]<br>90[70.9] | 100<br>[120.01] | 65<br>[120.01] | 65<br>[120.01] |

Table 7: Results for experiments on 13 real world networks. Average accuracy and time.

### Acknowledgements

This work was supported in part by the NSF grant IIS-0086529 and MURI ONR award N00014-00-1-0617.